\documentclass[runningheads]{llncs}
\usepackage{graphicx}
\usepackage{amsmath,amssymb} 
\usepackage{color}

\usepackage{mmstyle}
\usepackage{float}
\usepackage[dvipsnames]{xcolor}
\usepackage{subcaption}
\usepackage{multirow,makecell,footmisc,url}
\usepackage{gensymb}
\usepackage{sidecap}

\usepackage{wrapfig}
\usepackage{xspace}

\renewcommand{\etal}{\textit{et~al.}\@\xspace}

\newcommand*{\vs}{\textit{vs.}\@\xspace}

\usepackage{microtype}
\newcommand{\figref}[1]{Fig.~\ref{#1}}
\newcommand{\tblref}[1]{Table~\ref{#1}}
\newcommand{\sref}[1]{\S\ref{#1}}
	
\newcommand{\app}{\raise.17ex\hbox{$\scriptstyle\sim$}}

\newcolumntype{x}[1]{>{\centering\arraybackslash}p{#1pt}}
\newcommand{\dt}[1]{\fontsize{7pt}{0.1em}\selectfont (#1)}
\newlength\savewidth\newcommand\shline{\noalign{\global\savewidth\arrayrulewidth
		\global\arrayrulewidth 1pt}\hline\noalign{\global\arrayrulewidth\savewidth}}
\newcommand{\tablestyle}[2]{\setlength{\tabcolsep}{#1}\renewcommand{\arraystretch}{#2}\centering\footnotesize}

\makeatletter\renewcommand\paragraph{\@startsection{paragraph}{4}{\z@}
	{.5em \@plus1ex \@minus.2ex}{-.5em}{\normalfont\normalsize\bfseries}}\makeatother

\definecolor{citecolor}{RGB}{34,139,34}
\usepackage[pagebackref=true,breaklinks=true,letterpaper=true,colorlinks,
citecolor=citecolor,bookmarks=false]{hyperref}

\begin{document}
\pagestyle{headings}
\mainmatter
\def\ECCVSubNumber{***}  

\title{Feature Pyramid Grids\vspace{-5pt}} 

\titlerunning{Feature Pyramid Grids} 
\authorrunning{Chen \etal} 
\author{
   \hspace*{-30pt} Kai Chen$^{1, 2*}$ ~ Yuhang Cao$^{2}$ ~ Chen Change Loy$^{3}$ ~ Dahua Lin$^{2}$ ~ Christoph Feichtenhofer$^{4}$ \hspace*{-30pt} 
}
\institute{
    $^{1}$SenseTime Research \quad
    $^{2}$The Chinese University of Hong Kong \quad \\\vspace*{2pt}
    $^{3}$Nanyang Technological University \quad
    $^{4}$Facebook AI Research (FAIR)\vspace*{-5pt}
}
\renewcommand*{\thefootnote}{\fnsymbol{footnote}}
\setcounter{footnote}{1}
\footnotetext{Work done during an internship at Facebook AI Research}
\renewcommand*{\thefootnote}{\arabic{footnote}}
\setcounter{footnote}{0}

\maketitle

\begin{abstract}

Feature pyramid networks have been widely adopted in the object detection literature to improve feature representations for better handling of variations in scale. In this paper, we present Feature Pyramid Grids (FPG), a deep multi-pathway feature pyramid, that represents the feature scale-space as a regular grid of parallel  bottom-up pathways which are fused by multi-directional lateral connections. FPG can improve single-pathway feature pyramid networks by significantly increasing its performance at similar computation cost, highlighting importance of deep pyramid representations. In addition to its general and uniform structure, over complicated structures that have been found with neural architecture search, it also compares favorably against such approaches without relying on search. We hope that FPG with its uniform and effective nature can serve as a strong component for future work in object recognition.

\end{abstract}

\vspace{-10pt}

\section{Introduction}
\label{sec:intro}

It seems trivial how human perception can simultaneously recognize visual information across various levels of different resolution. 
For machine perception, recognizing objects at various scales has been a classical challenge in visual recognition over decades \cite{burt1981segmentation,burt1983laplacian,koenderink1984structure,mallat1989theory,perona1990scale}. Numerous methods have been developed to build pyramid representations \cite{burt1981segmentation,burt1983laplacian,koenderink1984structure} as an effective way to model the scale-space, by building a hierarchical pyramid ranging from large to small image scales. Such classical pyramid representations are typically built by subsequent filtering (blurring) and subsampling operations applied to the image.

In recent deep learning approaches, a bottom-up pathway is inherently built by ConvNets \cite{lecun1989backpropagation} that hierarchically abstract information from higher to lower resolution in deeper layers, also by hierarchical filtering and subsampling. For object detection tasks, Feature Pyramid Networks (FPN)~\cite{lin2017feature}, an effective representation for multi-scale features has become popular. FPN augments ConvNets with a second top-down pathway and lateral connections to enrich high-resolution features with semantic information from deeper lower-resolution features.

\definecolor{xucolor}{RGB}{160, 94, 148}
\newcommand{\xucolor}[1]{\textcolor{xucolor}{#1}}
\definecolor{xdcolor}{RGB}{169, 209, 142}
\newcommand{\xdcolor}[1]{\textcolor{xdcolor}{#1}}
\definecolor{xlcolor}{RGB}{68, 114, 169}
\newcommand{\xlcolor}[1]{\textcolor{xlcolor}{#1}}
\definecolor{xscolor}{RGB}{244, 177, 131}
\newcommand{\xscolor}[1]{\textcolor{xscolor}{#1}}
\definecolor{pathwaycolor}{RGB}{121,178,128}
\newcommand{\pathwaycolor}[1]{\textcolor{pathwaycolor}{#1}}

\newcommand{\backbonecolor}[1]{\textcolor{backbonecolor}{#1}}
\definecolor{backbonecolor}{RGB}{165,170,243}

\begin{figure}
	\centering
	\vspace{-15pt}
	\hspace*{-5pt}
	\includegraphics[width=0.95\linewidth]{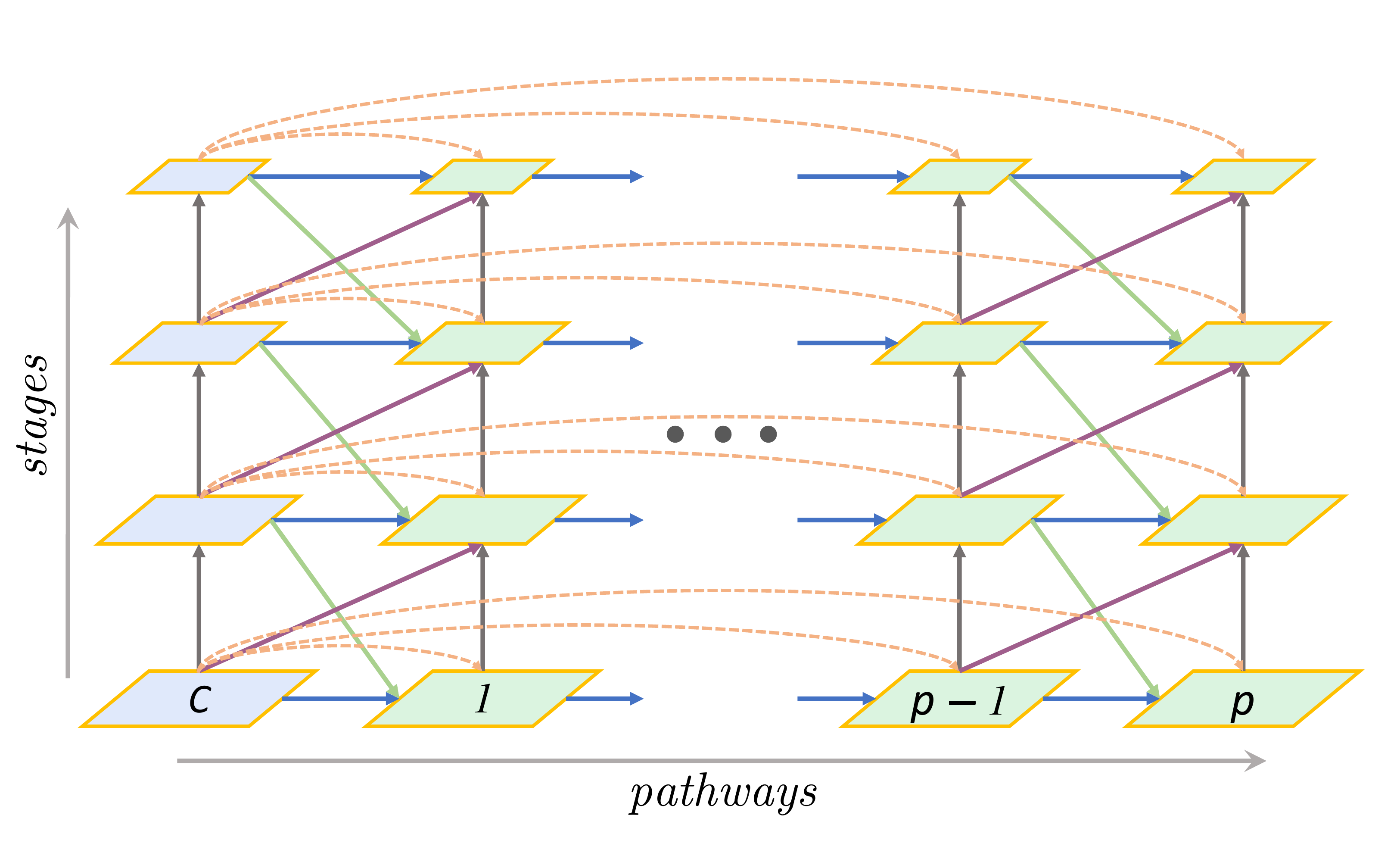}
		\vspace{-15pt}
	\caption{\textbf{A Feature Pyramid Grid (FPG)} connects the \textit{\backbonecolor{\textbf{backbone features}}}, $C$, of a ConvNet with a regular structure of $p$ parallel top-down pyramid \textit{\pathwaycolor{\textbf{pathways}}} which are fused by multi-directional \textit{lateral connections}, \mbox{ AcrossSame  \xlcolor{$\to$}}, AcrossUp  \xucolor{$\nearrow$},  AcrossDown \xdcolor{$\searrow$}, and AcrossSkip \xscolor{$\curvearrowright$}. AcrossSkip are direct connections while all other types use convolutional and ReLU layers.}
	\label{fig:teaser}
	\vspace{-15pt}
\end{figure}

There exist recent efforts~\cite{ghiasi2019fpn,xu2019auto}  of applying Neural Architecture Search (NAS) to find deeper feature pyramid representations that have shown strong experimental results. 
NAS-FPN~\cite{ghiasi2019fpn} defines a search
space for the modular pyramidal architecture and adopts reinforcement learning
to search the best performing one and Auto-FPN~\cite{xu2019auto} proposed new search spaces for both FPN and the box head.

While search-based approaches have indeed shown new levels of performance that outperform conventional, manually
designed FPN structures, there are several implications \wrt NAS in the context of finding improved architectures.

(i) The final network structure is often complicated hence not very comprehensible, limiting the adoption of the models by the community.

(ii) The search cost incurred by NAS is large, implicating up to thousands of TPU hours~\cite{ghiasi2019fpn} to find an optimal architecture.

(iii) The discovered architecture may not generalize well to other detection frameworks.
To give an example, \mbox{NAS-FPN} achieves a good performance on RetinaNet (which it was searched for), but it is unclear if it also performs similarly on other detection architectures.

In this paper, we present Feature Pyramid Grids (FPG), a deep multi-pathway feature pyramid network that represents the feature scale-space as a regular grid of parallel pathways fused by multi-directional lateral connections between them, as shown in  \figref{fig:teaser}.
FPG enriches the hierarchical \backbonecolor{feature representation} built internally in the backbone pathway of a ConvNet with \textit{multiple} pyramid \pathwaycolor{pathways} in parallel. On a high-level, FPG is a deep generalization of FPN \cite{lin2017feature} from one to $p$ pathways under a dense lateral connectivity structure.  

Different from FPN, all the individual pathways are built in a bottom-up manner, similar to the backbone pathway that goes from the input image to a prediction output.
To form a deep \textit{grid of feature pyramids}, the pyramid pathways are intertwined with various lateral connections, both across-scale as well as within-scale to enable  information exchange across all levels. We categorize these \textit{lateral connections} into four types, AcrossSame \xlcolor{$\to$}, AcrossUp \xucolor{$\nearrow$}, AcrossDown \xdcolor{$\searrow$}, and AcrossSkip \xscolor{$\curvearrowright$}, as illustrated in \figref{fig:teaser}.

Conceptually, the idea of FPG is generic, and it can be instantiated with different pathway and lateral connection design as well as  implementation specifics. 
Our experiments systematically analyze the importance of key components (parallel pathways and lateral connections) in feature pyramid design using our unified FPG framework. Our aim is to strike a good trade-off between accuracy and computation, and we approach it by starting with a densely connected pyramid grid that is subsequently contracted based on systematic ablation. 

In our evaluation, we are interested in the questions (1) if FPG can improve over FPN  under \textit{similar} complexity cost, and (2) if FPG could compete with NAS-optimized pyramid structures, even though being systematically designed. 

We apply FPG to single-stage (RetinaNet~\cite{lin2017focal}),
two-stage (Faster R-CNN~\cite{ren2015faster}, Mask R-CNN~\cite{he2017mask} and cascaded (Cascade R-CNN~\cite{cai2018cascade}) detectors.
Our findings are that, under similar computational cost, FPG performs better than FPN and NAS-FPN on various models and settings.
Concretely, adopting the same setting as in NAS-FPN, our FPG achieves 0.2\%, 1.5\%, 2.3\% and
2.2\% higher mean Average Precision (mAP) than NAS-FPN on RetinaNet, Faster R-CNN, Mask R-CNN and
Cascade R-CNN, respectively.

Our ablations reveal that a straightforward extension of FPN from one to many pathways does not succeed, but FPG is able to consistently increase accuracy for deeper pyramid representations. Overall, our experiments show that FPG is efficient and generalizes well across detectors, providing better performance than architecture searched pyramid networks. Given its systematic nature, we hope that FPG can serve as a strong component for future work in object recognition.

\section{Related Work}
\label{sec:related}

\paragraph{Handcrafted FPN architectures.}
Scale variation is a well-known challenge for instance-level recognition tasks, and building pyramidal representations have been an effective way to process visual information across various image resolutions, in classical computer vision applications \cite{burt1981segmentation,burt1983laplacian,koenderink1984structure,perona1990scale}, and also in deep learning based approaches \cite{liu2016ssd,cai2016unified,lin2017feature,kong2017ron,huang2017multi,liu2018path,yu2018deep,zhao2019m2det}. 

SSD~\cite{liu2016ssd} and MS-CNN~\cite{cai2016unified} utilize multi-level feature maps
to make predictions, but no aggregation is performed across different feature levels.
FPN~\cite{lin2017feature} is the current leading paradigm for learning feature
representation of different levels through the top-down pathway and lateral connections.
Similarly, RON~\cite{kong2017ron} introduces reverse connections to pass information
from high-level to low-level features.
Although MSDNet~\cite{huang2017multi} is not designed for FPN, it maintains
coarse and fine level features throughout the network with a two-dimensional multi-scale network architecture.
HRNet~\cite{sun2019deep} also maintains high-resolution representations through the whole backbone feedforward process.
PANet~\cite{liu2018path} extends FPN by introducing an extra bottom-up pathway to boost information flow.
DLA~\cite{yu2018deep} further deepens the representation by nonlinear and progressive fusion.
M2Det~\cite{zhao2019m2det} employs multiple U-Net to pursue a more suitable feature representation for object detection.

In relation to these previous efforts, FPG tries to formalize a unified grid structure that could potentially generalize many of these FPN variations within a systematic multi-pathway structure.

\paragraph{NAS-based FPN architectures.}
NAS automatically searches for efficient and effective architectures on a
specific task. It shows promising results on image classification~\cite{real2019regularized,liu2018darts,tan2019efficientnet},
and is also applied to other downstream tasks~\cite{chen2019detnas,peng2019efficient,liu2019training,ghiasi2019fpn,xu2019auto,liu2019auto}.
Some methods aim at discovering better FPN architectures.
NAS-FPN~\cite{ghiasi2019fpn} searches the construction of merging cells, \ie,
how to merge features at different scales.
It achieves significant accuracyf improvements with a highly complicated wiring pattern (\ie architecture).
Auto-FPN~\cite{xu2019auto} searches the architecture of both the FPN and head.
It defines a fully connected search space with various dilated convolution
operations, resulting in a more lightweight solution than NAS-FPN.
Unlike NAS-FPN or FPG, the pathways in Auto-FPN is fixed and the
connections of different stacks are not the same, thus making it not easily scalable.

In contrast to NAS-based FPN architectures, FPG can be seen as a more unified approach to feature pyramid representations, which is simple, intuitive
and easy to extend.
\section{Feature Pyramid Grids}
\label{sec:method}

Our objective in this paper is to design a unified and general multi-pathway feature pyramid representation. We aim to use the hierarchical feature representation built internally by ConvNets and enrich it with multiple pathways and lateral connections between them, to form a regular \emph{Feature Pyramid Grid} (FPG). The concept is illustrated in \figref{fig:teaser}.

Our generic grid has a backbone pathway (\sref{sec:backbone}) and multiple pyramid pathways (\sref{sec:grid}), which are fused by lateral connections (\sref{sec:lateral}) to define FPG.

\subsection{Backbone pathway}\label{sec:backbone} 

The backbone pathway can be the hierarchical feature representation of any ConvNet for image classification (\eg, \cite{Krizhevsky2012,Simonyan2015,Szegedy2015,He2016}).
This pathway is identical to what is used as the bottom-up pathway in FPN~\cite{lin2017feature}.
It has feature maps of progressively smaller scales from the input image to the output.
As in~\cite{He2016,lin2017feature}, feature tensors with the same scale belong to a network stage and the last feature map of each stage is denoted as $C_i$, where $i$ corresponds to the stage within the backbone hierarchy.
The spatial stride of feature tensors \wrt the input increases from early to deeper stages, as is common in image classification \cite{Krizhevsky2012,Simonyan2015,Szegedy2015,He2016}.

\subsection{Pyramid pathways}\label{sec:grid}

The deeper backbone stages, closer to the classification layer of the network represent high-level semantics, but at low resolution, while the features in early stages are only weakly related to semantics,
but, on the other hand, have high localization accuracy due to their fine resolution.
The objective of the pyramid pathways is to build fine resolution features with strong semantic information.
A single pyramid pathway  consecutively upsamples deeper features of lower resolution to higher resolution of early stages,
aiming to propagate semantic information backwards towards the network input, in parallel to the backbone (\ie, feedforward) pathway.

\paragraph{Multiple pyramid pathways.}
FPG extends this idea by having \textit{multiple}, $p > 1$, pyramid pathways in parallel.
Our intention here is to enrich the capacity of the network to build a powerful representation with fine resolution across spatial dimensions and high discriminative ability, by employing multiple pyramid pathways in parallel.
A typical value is $p=9$ parallel pathways in our experiments.
We build the pyramid pathways in a \textit{bottom-up} manner, in parallel to the backbone pathway (and the first highest resolution pyramid feature is taken from the corresponding backbone stage).
Connections in pyramid pathways are denoted as \emph{SameUp}.
The presence of multiple pathways is key to the FPG concept (\figref{fig:teaser}) since it allows the network to build stronger pyramid features as will be demonstrated in our experiments. To form a deep Feature Pyramid Grid, the $p$ individual pyramid pathways are intertwined with various lateral connections introduced in the next section. 

\paragraph{Low channel capacity.}
Following the efficient design of FPN~\cite{lin2017feature}, we aim to make the pyramid pathways lightweight by reducing their channel capacity.
Concretely, the pyramid pathways use a significantly lower channel capacity than the number of channels of the final stage in the backbone pathway.
The typical value is $256$ in FPN. Notice that the computation cost (floating-number operations, or FLOPs) of a weight layer scales \emph{quadratically} with its channel dimensions (\ie width). Therefore, reducing the channel capacity in the pyramid pathways can make multiple pathways very computationally-effective as we will demonstrate in our experiments.

\subsection{Lateral connections}\label{sec:lateral}
The aim of lateral connections is to enrich features with multi-directional (semantic) information flow in the scale space, and allow complex hierarchical feature learning across different scales.

We are using across-scale as well as within-scale connections between adjacent pathways. 
In relation to this, our $p$ parallel pyramid pathways with the lateral connections between define a Feature Pyramid Grid. 

We categorize our lateral connections into 4 different categories according
to their starting and ending feature stages, which are denoted as:
\begin{itemize}
	\setlength\itemsep{.2em}
\item Across-pathway same-stage (\emph{AcrossSame}, \xlcolor{$\to$})
\item Across-pathway bottom-up connection (\emph{AcrossUp}, \xucolor{$\nearrow$})
\item Across-pathway top-down (\emph{AcrossDown},  \xdcolor{$\searrow$})
\item Across-pathway skip connection (\emph{AcrossSkip}, \xscolor{$\curvearrowright$})
\end{itemize}

We describe how these connections are implemented within the context of a concrete instantiation of FPG next.

\subsection{Instantiations}\label{sec:realization}

Our idea of FPG is generic, and it can be instantiated with different pathway and lateral connection designs as well as  implementation specifics. Here, we describe concrete instantiations of the FPG network architectures.

\paragraph{Backbone pathway.}
The backbone pathway is the feedforward computation of the backbone ConvNet, which computes a feature hierarchy consisting of feature maps at several scales with a scaling step of 2 (\ie the spatial stride between stages).
Taking ResNet~\cite{He2016} for example, we adopt the same scheme as in FPN and use the output feature map of each stage's last residual block to represent the pyramid levels, denoted as $\{C_2, C_3, C_4, C_5\}$.

\paragraph{Pyramid pathways.}
Similar to the backbone pathway, pyramid pathways represent information across scales. 
We follow a simple design for building these in a bottom-up manner, starting from the highest resolution stage to the lowest. The first feature map of the pathway is formed by a $1\times 1$  lateral connection from the corresponding high-resolution backbone or pyramid stage.
Then, we use sub-sampling to create each lower-level feature map in the pyramid pathway by using a $3\times 3$ convolution width stride 2, Therefore, in each pyramid pathway, the feature hierarchy consists of multi-scale feature maps with the same spatial resolution of the individual stages as in the backbone pathway.

\paragraph{Lateral connections.}
Our lateral connections fuse between the pathways into multiple directions.
We employ across-pathway lateral, bottom-up, and top-down connections between adjacent pathways,
and skip connections between the first pathway and all other pathways.
Concrete instantiations of the lateral connections are as follows.
\begin{itemize}
\setlength\itemsep{.2em}
\item \emph{AcrossSame}, \xlcolor{$\to$}

These lateral connections to connect the same-level features across pathways.
We use a $1\times1$ lateral convolution on each feature map to project the features and fuse them with the corresponding features in the adjacent pathway.

\item \emph{AcrossUp}, \xucolor{$\nearrow$}

In order to shorten the path from low-level features in shallow pathways to
high-level features in deep pathways,
we introduce direct connections to build the across-level bottom-up pathway.
The low-level feature map is downsampled to half size by a $3\times 3$ stride-2 convolution
and then fused with the higher-level one.

\item \emph{AcrossDown},  \xdcolor{$\searrow$}

Similar to our bottom-up information stream within each pathway, we aim for a top-down flow of information by incorporating \emph{AcrossDown} connections. 
Firstly we upsample the high-level feature maps by a scaling factor of 2 with nearest interpolation,
and then use a $3\times3$ convolution to make $AcrossDown$ learnable.
The upsampled features are fused with the low-level features.

\item \emph{AcrossSkip}, \xscolor{$\curvearrowright$}

To ease the training of such a wide feature pyramid grid,
we add skip connections, \eg, $1\times1$ convolution, between same level of the first pathway and each later pathway.

\end{itemize}

Identically as for building the $p$ parallel pathways, we employ element-wise \textit{Sum} as the fusion function for all lateral connections. 

\subsection{FPG implementation details} \label{sec:grid_details}

Given a feature hierarchy in the backbone pathway, \eg, $\{C_2,C_3,C_4,C_5\}$
with strides of $\{4, 8, 16, 32\}$ respectively, as in FPN, we first use $1\times1$ convolutions to uniformly reduce the channel capacity by $\beta$ times the width of the highest feature map in the backbone pathway (\eg, 256 for a ResNet with a maximum of width of 2048 and  $\beta=1/8$), producing $\{P_2^1,P_3^1,P_4^1,P_5^1\}$.
Similar to FCOS~\cite{tian2019fcos}, we produce $P_6^1$ from $P_5^1$.
Then we apply the same topology to each following pathway, until the last one $\{P_2^p,P_3^p,P_4^p,P_5^p,P_6^p\}$.
We follow the standard approach of using $P_2\sim P_6$ for Faster R-CNN and Mask R-CNN, and $P_3\sim P_7$ for \mbox{RetinaNet}~\cite{lin2017focal}, everything else is identical across detectors.

Following~\cite{ghiasi2019fpn}, each convolution block in above lateral connections consists of a ReLU \cite{Krizhevsky2012},
a convolution layer, and a BatchNorm \cite{Ioffe2015} layer.
Those connections are not shared across different pathways.
After the last pathway, we append a vanilla $3\times3$ convolution layer
after each merged feature map to output the final feature map.

We aim for the simplest possible design of FPG.  
We think that adopting advanced upsampling and downsampling operators such as~\cite{wang2019carafe,gao2019lip}, separable-convolution~\cite{Howard2017,xu2019auto}, designing more advanced blocks or fusion strategies (\eg using attention \cite{hu2018squeeze,ghiasi2019fpn}), may further boost
the system-level performance, but is not the focus of this work.

\begin{figure}
	\centering
			\vspace{-5pt}
	\includegraphics[width=0.85\linewidth]{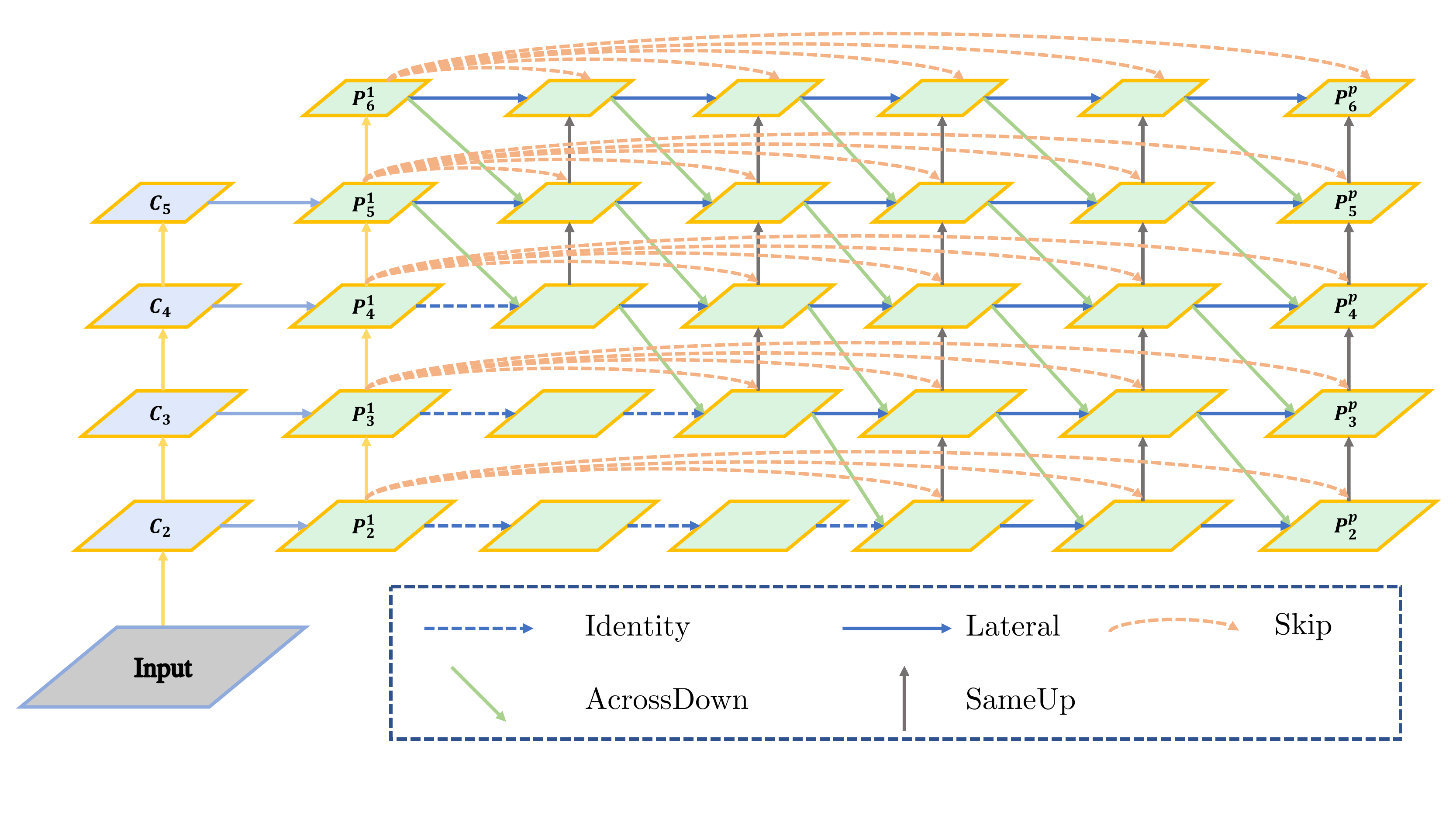}
		\vspace{-20pt}
	\caption{ The contracted FPG, in which \emph{AcrossUp} connections are ablated and the lower triangular connections are truncated. This conserves computation while preserving accuracy (\sref{subsec:ablation}). The illustration shows 5/9 pathways of the grid.
	}
	\label{fig:fpg}
	\vspace{-10pt}
\end{figure}

\paragraph{Grid contraction.}
Our ablation studies in \sref{subsec:ablation} reveal that the regular design of FPG can be simplified for better computation/accuracy trade-off. This will be demonstrated in our experiments, but for now, we show a more efficient version of FPG that reduces some stages without sacrificing significant accuracy.
First, there are two bottom-up streams in our design: \emph{SameUp} and \emph{AcrossUp}. Our ablation analysis in \sref{subsec:ablation} reveals that removing \emph{AcrossUp} has no significant impact and \emph{SameUp} is sufficient to provide the information flow from low-level to high-level features  which is expected to be less rich in semantic information, and therefore might require lower representation capacity.
Second, we found that contracting the ``lower triangle'' connections for high resolution feature maps can be done without sacrificing performance. 
Our hypothesis is that low-level feature maps need first to be enriched from top-down propagation before benefiting from deep pathways structures. 
Specifically, the lower triangular part of the grid can be truncated to conserve computation while preserving accuracy. The contracted FPG architecture is illustrated in \figref{fig:fpg} and used by default in the experiments.

\section{Experiments}
\label{sec:experiments}
\vspace{-5pt}

We perform experiments on standard object recognition tasks of object detection and instance segmentation.
We compare to FPN~\cite{lin2017feature} and  NAS-FPN \cite{ghiasi2019fpn} as they represent the closest related shallow and deep feature pyramid networks, respectively. Comparisons to other other pyramid networks (PANet~\cite{liu2018path},  HRNet~\cite{sun2019deep}) are provided in appendix \ref{app:comparison}.

\vspace{-5pt}
\subsection{Experimental Setup} \label{sec:exp_setup}
\vspace{-5pt}
\paragraph{Dataset and evaluation metric.}
We conduct experiments on the MSCOCO 2017 dataset~\cite{lin2014microsoft}.
For all tasks, models are trained on the \texttt{train} split and results are
reported on \texttt{val} splits.
We also show our main results on \texttt{test-dev}.
Evaluation follows standard COCO mAP metrics~\cite{lin2014microsoft}.

\paragraph{Implementation details.}

For object detection and instance segmentation, we experiment with two
different augmentation settings, denoted as \texttt{crop-aug} and \texttt{std-aug}.
\texttt{crop-aug} is the setting introduced in NAS-FPN \cite{ghiasi2019fpn}, images are firstly rescaled with a ratio randomly selected from the range [0.8, 1.2] and then cropped to a fixed size of $640\times640$.
We utilize 8 GPUs for training and use a batch-size of 8 images on each GPU, resulting in a total batch size of 64.
The training lasts for 50 epochs. The  initial learning rate is set to 0.08,
and then decays by 0.1 after 30 and 40 epochs.
BN layers are not frozen in the Pyramid but frozen in other components.
\texttt{std-aug} is the standard augmentation procedure in the original publications of the detectors used \cite{ren2015faster,he2017mask,lin2017focal,cai2018cascade}. It resizes
input images to a maximum size of $1333\times800$ without changing the aspect ratio. Therefore it requires more GPU memory for training.
We use 16 GPUs for training and  the mini-batch size is 1 image per GPU (so the total mini-batch size is 16).
Models are trained for 12 epochs with an initial learning rate of 0.02
and decreased by 0.1 after 8 and 11 epochs.
BN statistics are synchronized across GPUs for the Pyramid, and frozen in the ImageNet pre-trained backbone, pathways.

We use ResNet-50 for the \texttt{crop-aug} setting experiments and ResNet-50/101 for the \texttt{std-aug} setting.
We use a weight decay of 0.0001 and momentum of 0.9.

For inference we exactly follow the standard settings in the original architectures \cite{ren2015faster,he2017mask,lin2017focal,cai2018cascade,ghiasi2019fpn}. For the \texttt{crop-aug} setting, the longer side of the images is resized to 640 pixels, while preserving aspect ratio \cite{ghiasi2019fpn}.
For the \texttt{std-aug} setting, we resize the images to a maximum size of $1333\times 800$ without changing the aspect ratio \cite{he2017mask}.
For object detection, the number of RoI (Region of Interest) proposals is 1000 for Faster R-CNN. The box prediction branch is applied on the proposals, followed by non-maximum suppression \cite{Girshick2015} with an IoU threshold of 0.5. We keep at most 100 bounding boxes for each image after NMS. For instance segmentation, the mask branch is run on the 100 highest scoring detection boxes \cite{he2017mask}. As in Mask R-CNN \cite{he2017mask}, the mask output is resized to the RoI size, and binarized at a threshold of 0.5.

We denote the combination of number of pathways $p$ and the common pathway channel width $w$  as $p$@$w$.
For example, using this terminology 9@256 indicates 9 pathways of channel width 256 for all pyramid layers.

We report single image floating point operations (FLOPs) as a basic \emph{unit} of measuring computational cost agnostic to implementation and hardware specifics. We are also showing number of overall parameters of the detection systems used.

\paragraph{Pyramid details.}
We apply FPG to various detectors with the \texttt{crop-aug} and \texttt{std-aug} settings.
We report the performance of 2 different architectures: \textit{FPG (9@256)} and \textit{FPG (9@128)}.
FPG (9@256) has comparable FLOPs with NAS-FPN (7@256), and FPG (9@128) is as
lightweight version of FPG that roughly matches the computational cost of  the default FPN (1@256).

\subsection{Main Results on Object Detection}
\begin{table}[t]
	    \fontsize{8pt}{1em}\selectfont
	\centering
	\caption{\small\textbf{Object detection} mAP on COCO \texttt{test-dev}.  Results of different detectors on COCO with the \emph{crop-aug} setting and inference FLOPs are reported on a single image of size 640x640. Backbone: ResNet-50 \cite{He2016}. FPG achieves significant accuracy gains at similar complexity.}
	\vspace{5pt}
	\begin{tabular}{ l | l | c c l | c c c c c c}
		Detector & Pyramid & FLOPs & Params & AP & $\text{AP}_{50}$ & $\text{AP}_{75}$ & $\text{AP}_{S}$ & $\text{AP}_{M}$ & $\text{AP}_{L}$ \\
		\shline
		\multirowcell{4}{RetinaNet} & FPN 1@256      & 95.7 & 37.8  & 37.0 & 55.9 & 39.7 & 16.1 & 41.1 & 51.5 \\
		& \textbf{FPG} 9@128 & 95.9  & 40.1  & \textbf{39.0\dt{+2.0}} & 58.2 & 41.9 & 17.3 & 43.9 & 53.3 \\
		\cline{2-10}
		& NAS-FPN 7@256  & 138.8 & 59.8 & 39.8 & 58.5 & 42.6 & 17.6 & 44.8 & 54.4 \\
		& \textbf{FPG} 9@256  & 136.0 & 72.5 & \textbf{40.0}\dt{+0.2} & 59.1 & 42.9 & 18.2 & 45.0 & 54.7 \\
		\hline
		\multirowcell{4}{Faster\\ R-CNN } & FPN 1@256    & 91.4  & 41.5 & 37.6 & 58.4 & 40.7 & 18.4 & 40.7 & 50.8 \\
		& \textbf{FPG} 9@128  & 99.1  & 42.1  & \textbf{40.0\dt{+2.4}} & 59.9 & 43.5 & 20.0 & 43.6 & 53.8 \\\cline{2-10}
		& NAS-FPN 7@256    & 265.3 & 68.2  & 39.9 & 58.8 & 43.3 & 18.8 & 43.8 & 54.4 \\
		& \textbf{FPG} 9@256  & 254.1 & 79.8  & \textbf{41.4\dt{+1.5}} & 61.4 & 45.1 & 21.5 & 44.8 & 54.8 \\
		\hline
		\multirowcell{4}{Mask\\ R-CNN} & FPN 1@256      & 159.9 & 44.2  & 38.6 & 59.2 & 41.9 & 18.7 & 41.4 & 52.4 \\
		& \textbf{FPG} 9@128    & 161.8 & 44.4  & \textbf{40.9\dt{+2.3}} & 60.5 & 44.6 & 20.9 & 44.4 & 54.6 \\\cline{2-10}
		& NAS-FPN 7@256      & 333.8 & 70.8  & 40.1 & 57.9 & 44.3 & 19.0 & 45.7 & 58.1 \\
		& \textbf{FPG} 9@256    & 322.6 & 82.4 & \textbf{42.4\dt{+2.3}} & 62.1 & 46.3 & 22.5 & 45.8 & 56.0 \\
		\hline
		\multirowcell{4}{Cascade\\ R-CNN} & FPN 1@256      & 119.1 & 69.2   & 40.6 & 58.5 & 43.9 & 19.5 & 43.4 & 55.5 \\
		& \textbf{FPG} 9@128 & 113.9 & 56.9   & \textbf{42.5\dt{+1.9}} & 60.0 & 46.0 & 21.4 & 45.9 & 57.3 \\\cline{2-10}
		& NAS-FPN 7@256  & 292.9 & 95.8   & 41.6 & 58.3 & 45.1 & 19.1 & 45.8 & 57.3 \\
		& \textbf{FPG} 9@256 & 281.8 & 107.4  & \textbf{43.8\dt{+2.2}} & 61.5 & 47.6 & 23.2 & 47.2 & 58.2 \\
	\end{tabular}
	\label{tab:results1}
	\vspace{-15pt}
\end{table}

The results of the \texttt{crop-aug} setting are shown in Table~\ref{tab:results1}, for four different detection architectures.
For all detection systems, \ie, single-stage (RetinaNet)~\cite{lin2017focal},
two-stage (Faster R-CNN, Mask R-CNN)~\cite{ren2015faster,he2017mask} and cascaded (Cascade R-CNN)~\cite{cai2018cascade}, \textbf{FPG} outperforms FPN~\cite{lin2017feature} by a strong margin, and also achieves better performance than NAS-FPN~\cite{ghiasi2019fpn} , without relying on architecture search.

More specifically, compared to FPN, the \textbf{FPG} (9@128) improves the box AP by \textbf{+2.0}, \textbf{+2.4}, \textbf{+2.3},
and \textbf{+1.9} mAP  on RetinaNet, Faster R-CNN, Mask R-CNN, and Cascade R-CNN, respectively. This result is achieved at roughly the same computational cost (FLOPs), showing that deeper and densely connected feature pyramid representations ({FPG}), that are thinner in width (128 \vs 256 channels), are significantly better than wider but shallower ones in FPN.

Compared with NAS-FPN (7@256), \textbf{FPG} (9@256) achieves +0.2, \textbf{+1.5}, \textbf{+2.3} and \textbf{+2.2}
higher mAP on those four detectors, {while maintaining slightly less FLOPs}.
Without any bells and whistles, FPG (9@256) obtains an mAP of 41.4\% on Faster R-CNN
and 43.8\% on Cascade R-CNN using a ResNet-50 backbone, significantly better than NAS-FPN~\cite{ghiasi2019fpn} under \textit{identical} settings. 

Interestingly, NAS-FPN performs comparably to FPG on RetinaNet (which it was optimized for), but achieves inferior results on the other detection systems.
As we observe higher gains over NAS-FPN in multi-stage detectors, this  suggests that the NAS-FPN architecture found for the single-stage detection architecture (\ie RetinaNet) might not generalize well to multi-stage detectors.
Our systematic multi-pathway approach in FPG exhibits {good generalization across all detection systems with strong gains over NAS-FPN for Faster R-CNN, Mask R-CNN, and Cascade R-CNN, even though it does not benefit from architecture search.
The results of the \texttt{std-aug} setting are shown in Table~\ref{tab:results2} and show slightly lower gains, but are consistent with our findings for the \texttt{crop-aug} setting.

\begin{table}[h!]
	\fontsize{8pt}{1em}\selectfont
	\vspace{-15pt}
	\centering
	\caption{\textbf{Object detection} mAP on COCO \texttt{test-dev}.  Results of different detectors on COCO with the \emph{std-aug} setting and inference FLOPs are reported on a single image of size 1280x832.}
	\vspace{5pt}
	\addtolength{\tabcolsep}{-0.75pt}
	\begin{tabular}{ l | l | c c l | c c c c c c c}
		Detector & Pyramid & FLOPs & Params  & AP & $\text{AP}_{50}$ & $\text{AP}_{75}$ & $\text{AP}_{S}$ & $\text{AP}_{M}$ & $\text{AP}_{L}$ \\
		\shline
		\multirowcell{4}{Faster\\ R-CNN R-50} & FPN 1@256  & 214.8 & 41.5 & 36.6 & 58.8 & 39.6 & 21.6 & 39.8 & 45.0 \\
		& \textbf{FPG} 9@128                               & 245.0 & 42.1 & \textbf{38.0\dt{+1.4}} & 59.4 & 41.2 & 22.1 & 40.7 & 46.4 \\\cline{2-10}
		& NAS-FPN 7@256                                    & 666.9 & 68.2 & 39.0 & 59.5 & 42.4 & 22.4 & 42.6 & 47.8 \\
		& \textbf{FPG} 9@256                               & 637.8 & 79.8 & \textbf{39.2}\dt{+0.2} & 60.8 & 42.7 & 22.7 & 41.9 & 48.4 \\
		\hline
		\multirowcell{4}{Faster\\ R-CNN R-101} & FPN 1@256  & 294.0 & 60.5 & 38.8 & 60.9 & 42.3 & 22.3 & 42.2 & 48.6 \\
		& \textbf{FPG} 9@128                                & 324.2 & 61.1 & \textbf{39.5}\dt{+0.7} & 61.0 & 43.0 & 22.9 & 42.4 & 49.2 \\\cline{2-10}
		& NAS-FPN 7@256                                     & 746.0 & 87.2 & 40.3  & 61.2 & 43.8 & 23.1 & 43.9 & 50.1 \\
		& \textbf{FPG} 9@256                                & 716.9 & 98.8 & \textbf{40.6}\dt{+0.3} & 62.2 & 44.3 & 23.4 & 43.5 & 50.6 \\
		\hline
		\multirowcell{4}{Mask\\ R-CNN R-50} & FPN 1@256 & 283.4 & 44.2  & 37.4 & 59.3 & 40.7 & 22.0 & 40.6 & 46.3 \\
		& \textbf{FPG} 9@128                            & 307.8 & 44.5  & \textbf{39.0\dt{+1.6}} & 59.9 & 42.4 & 22.8 & 41.8 & 48.4 \\\cline{2-10}
		& NAS-FPN 7@256                                 & 735.4 & 70.8  & 39.6 & 59.8 & 43.3 & 22.8 & 42.7 & 48.4 \\
		& \textbf{FPG} 9@256                            & 706.3 & 82.4  & \textbf{40.3}\dt{+0.7} & 61.2 & 44.2 & 23.7 & 42.8 & 49.7 \\
		\hline
		\multirowcell{4}{Mask\\ R-CNN R-101} & FPN 1@256 & 362.5 & 63.2  & 39.7 & 61.6 & 43.2 & 23.0 & 43.2 & 49.7 \\
		& \textbf{FPG} 9@128                             & 386.9 & 63.5  & \textbf{40.5}\dt{+0.7} & 61.5 & 44.3 & 23.5 & 43.6 & 50.2 \\\cline{2-10}
		& NAS-FPN 7@256                                  & 814.5 & 89.8  & 40.5   & 60.8 & 44.2 & 23.4 & 43.7 & 50.2 \\
		& \textbf{FPG} 9@256                             & 785.4 & 101.4 & \textbf{41.6\dt{+1.1}} & 62.7 & 45.5 & 24.1 & 44.5 & 51.6\\
	\end{tabular}
	\vspace{-15pt}
	\label{tab:results2}
\end{table}

\begin{wrapfigure}[16]{}{0.45\textwidth}
	\centering
	\vspace{-10pt}
	\includegraphics[width=0.5\textwidth]{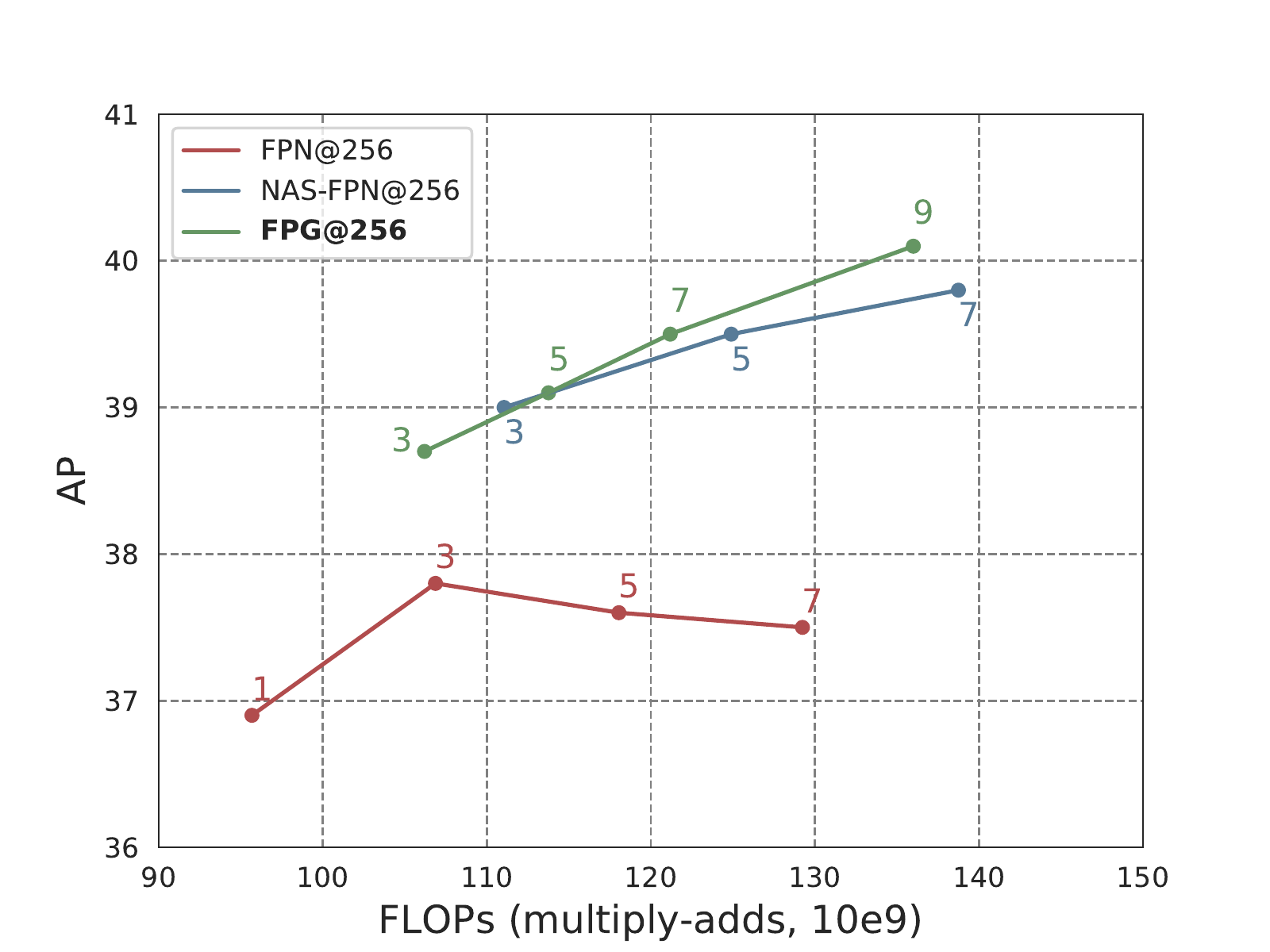}
	\vspace{-10pt}
	\caption{\small The efficiency-accuracy trade-off by increasing number of pathways. Extending FPN does not work beyond 3. Detector: RetinaNet (which NAS-FPN is searched on). Backbone: ResNet-50.}
	\label{fig:stack_vs_map}
\end{wrapfigure}
\paragraph{Efficiency \textit{vs.} accuracy trade-off.}

Feature pyramid architectures allow to  easily  configure the model capacity.
\begin{wrapfigure}[18]{}{0.47\textwidth}
	\centering
	\vspace{-15pt}
	\includegraphics[width=0.5\textwidth]{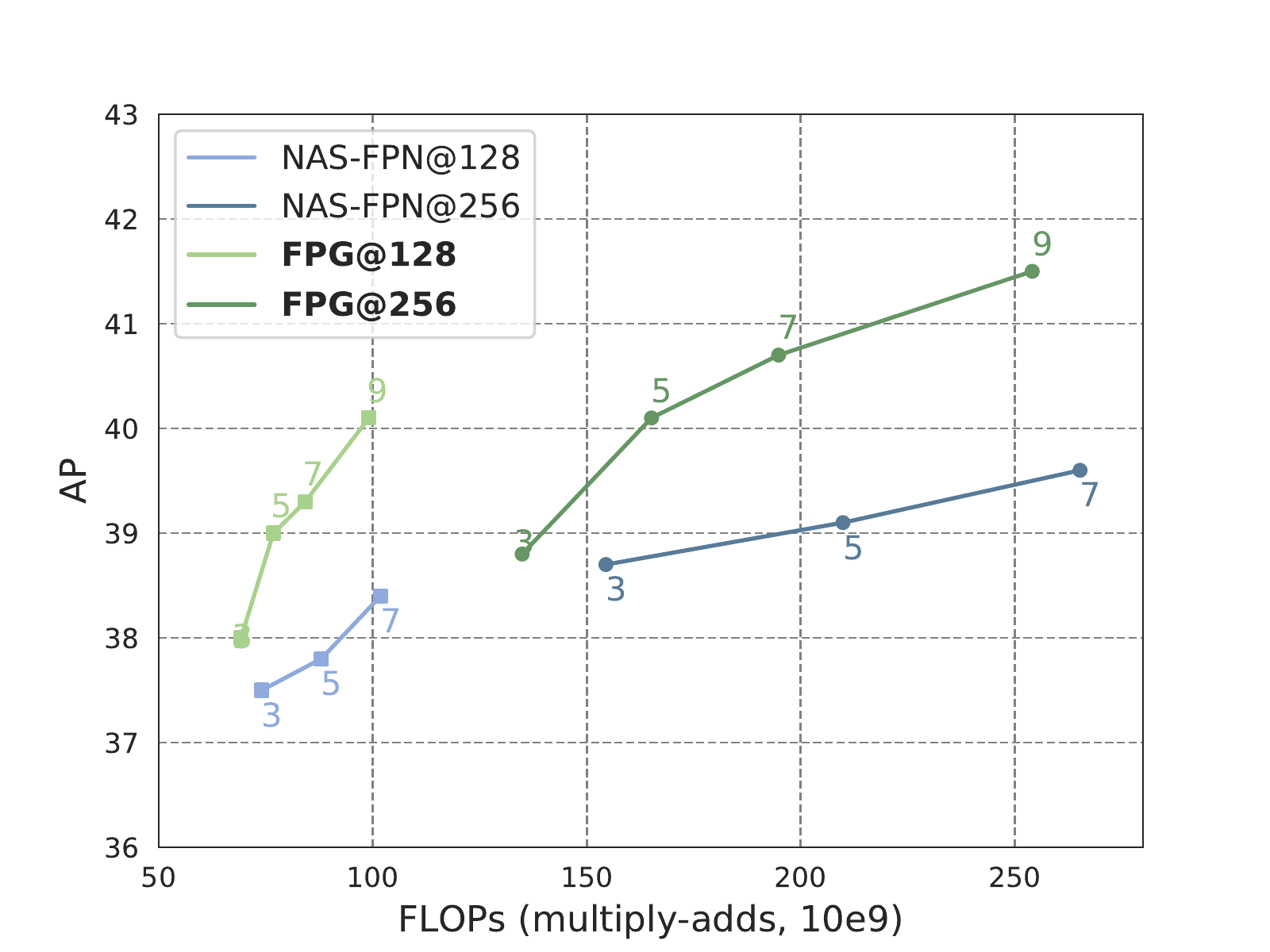}
	\caption{\small Efficiency-accuracy trade-off. FPG shows consistent improvement over NAS-FPN for varying pathway width (128/256) and depth (3, 5, 7, 9).
		The results are under the \texttt{crop-aug} setting and
		show the performance on COCO \emph{val2017}. Detector: Faster R-CNN. Backbone: ResNet-50.  }
	\label{fig:dim_vs_map}
\end{wrapfigure}  By adjusting the number of pyramid pathways $p$ (depth) and the pathway width $w$ we can trade off the efficiency with accuracy and obtain a set of FPG networks, from lightweight to heavy computational cost. 
We apply the same principle to FPN~\cite{lin2017feature}  and NAS-FPN~\cite{ghiasi2019fpn}, which also uses this strategy to stack capacity, and investigate the compute/accuracy trade-off.

Figure~\ref{fig:stack_vs_map} shows the comparison on \mbox{RetinaNet}, which NAS-FPN was optimized for. 
First we notice, that extending FPN from one to many pathways does not succeed. We observe that for extending FPN with more than 3 top-down pathways  accuracy stops increasing and it will instead decrease. This verifies that a simple extension of FPN to multiple top-down pathways does not lead to similar performance as we can achieve with FPG.
By changing the pyramid depth, we observe that increasing the FPG pyramid pathways or NAS-FPN stacks is beneficial in terms of accuracy (vertical axis, AP). Overall, FPG achieves a better trade-off than FPN and NAS-FPN. For example, FPG (9@256) achieves higher accuracy than NAS-FPN (7@256) with fewer FLOPs.

Next we investigate the Faster \mbox{R-CNN} detector and vary the number of pyramid pathways $p$ (depth) \textit{and} the pathway width $w$ for studying the computation/accuracy trade-off. 

Figure~\ref{fig:dim_vs_map} shows the effects of multiple (\ie 3, 5, 7, 9) FPG pyramid pathways, or  NAS-FPN stacks, as well as varying the channel width (128, 256).
For both NAS-FPN and FPG, adopting a larger channel width of 256 improves the
accuracy (vertical axis) while resulting in higher FLOPs (horizontal axis).
We observe that  FPG achieves significantly better efficiency-accuracy trade-off for
 Faster R-CNN detectors for which NAS-FPN was not searched (it was optimized on RetinaNet), illustrating better generalization of FPG across detectors.

\subsection{Main Results on Instance Segmentation}
\vspace{-10pt}
\begin{table}[h]
	\centering
		\fontsize{8pt}{1em}\selectfont
	\caption{\small \textbf{Instance segmentation}  \emph{mask} AP on COCO \texttt{val2017}. FPG provides significant improvements over the FPN and NAS-FPN variants. Backbone: R-50.}
	\vspace{5pt}
	\tablestyle{3pt}{1.05}
    \begin{tabular}{l|c|c|c|c|c|c}
		Pyramid & AP & AP$_\text{50}$ & AP$_{75}$ & AP$_\text{S}$ & AP$_\text{M}$ & AP$_{L}$\\
		\shline
		FPN     & 34.5	& 55.2	& 36.8	& 13.4	& 37.2	& 54.0 \\
		NAS-FPN & 35.6	& 55.2	& 38.5	& 13.1	& 39.3	& 56.9 \\ \hline
		\textbf{FPG}    & \textbf{37.2} & \textbf{58.4} & \textbf{39.8} & \textbf{15.9} & \textbf{40.3} & \textbf{57.0}\\
	\end{tabular}
	\vspace{-10pt}
	\label{tab:ins-seg}
\end{table}

Here, we compare  the instance segmentation results of Mask R-CNN in \tblref{tab:ins-seg}. The setting is the same as in Table~\ref{tab:results1}, but for mask- instead of box-level prediction. 
\textbf{FPG} (9@256) achieves \textbf{+2.7} higher mask AP than FPN and \textbf{+1.6} higher mask AP than NAS-FPN (7@256), demonstrating generalization of FPG across different tasks. In general, our results show that a systematically designed feature pyramid grid can rival (RetinaNet) or even surpass (Faster R-CNN, Mask R-CNN and Cascade R-CNN) neural architecture search based optimization. We show qualitative results in \figref{fig:ins_seg_example}, where we compare our FPG with FPN and NAS-FPN. The visualizations show that FPN and NAS-FPN are challenged by misclassification of overlapping instances, as well as small-scale objects. More examples are available in appendix~\ref{app:qualitative_results}.

\begin{figure}[tb]
	\includegraphics[width=1.0\linewidth]{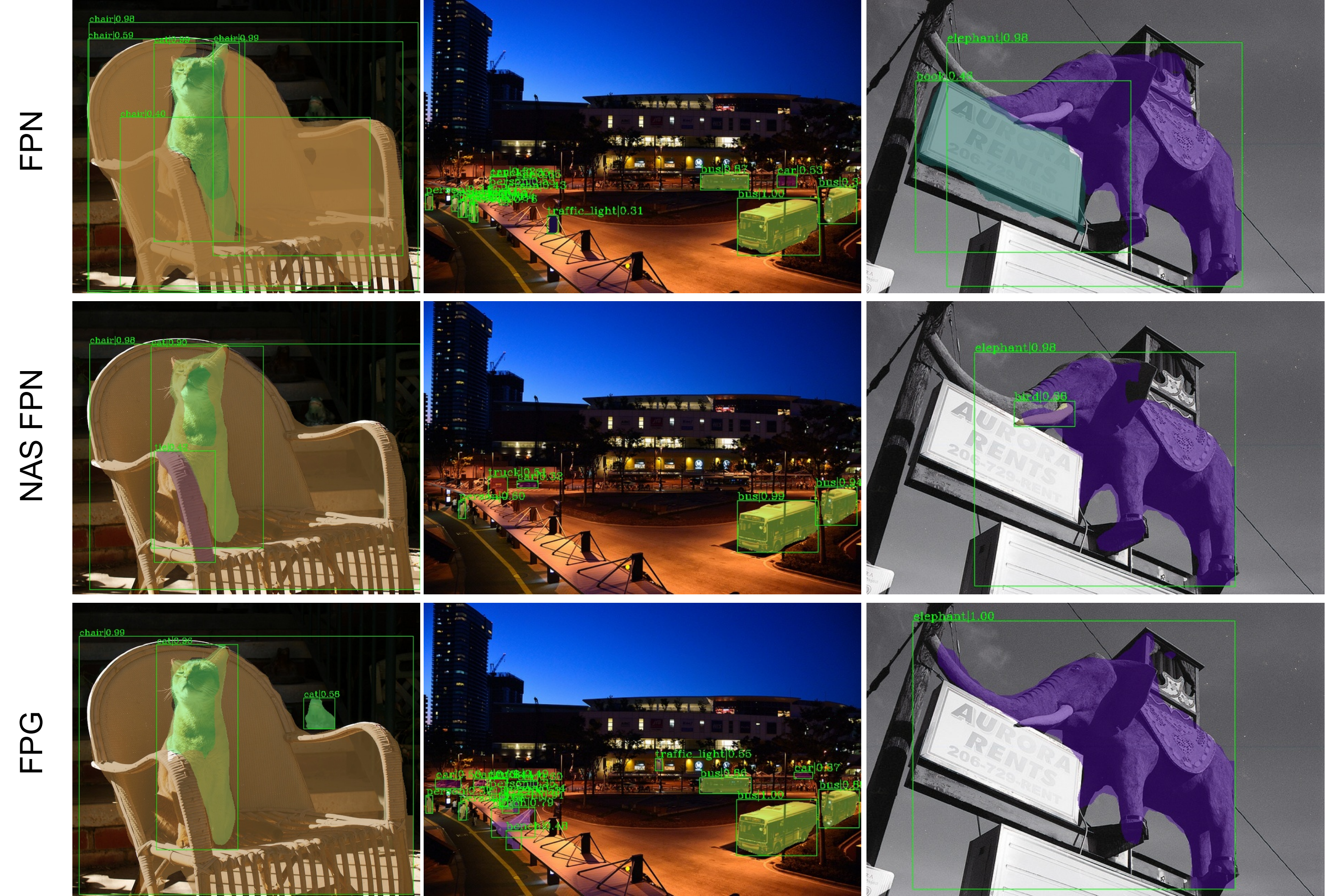}
\caption{\small \textbf{Example results for Instance Segmentation} with Mask R-CNN~\cite{he2017mask} using ResNet-50~\cite{He2016} with FPN~\cite{lin2017feature}, NAS-FPN~\cite{ghiasi2019fpn} and our FPG. Note how FPG is able to produce correct mask predictions for small-scale objects and has fewer misclassifications.}
	\vspace{-2mm}
	\label{fig:ins_seg_example}
\end{figure}

\begin{table}[htb]
	\centering
	\small
		\fontsize{8pt}{1em}\selectfont
	\vspace{-15pt}
	\caption{\small \textbf{Ablations}. We study the effectiveness of each component of FPG on \texttt{val2019} and report \emph{box} AP. Last row: Our default, contracted (Cont) instantiation.}
	\vspace{5pt}
	\tablestyle{3pt}{1.05}
	\begin{tabular}{ c c c c c | c c | c c c c c c}
		AD  & AU & SU & AS & Cont & FLOPs & Params & AP & AP$_\text{50}$ & AP$_{75}$ & AP$_\text{S}$ & AP$_\text{M}$ & AP$_{L}$\\
		\shline
		\checkmark & \checkmark & \checkmark & \checkmark & & 173.3 & 104.5 & \textbf{40.1} & 59.0 & 43.0 & 19.9 & 45.6 & 56.7 \\
		& \checkmark & \checkmark & \checkmark & & 128.1 & 83.3  & 35.5 & 52.9 & 38.2 & 14.0 & 39.7 & 53.7 \\
		\checkmark &            & \checkmark & \checkmark & & 162.0 & 83.3  & 40.1 & 59.0 & 42.9 & 19.6 & 46.1 & 56.4 \\
		\checkmark & \checkmark &            & \checkmark & & 162.0 & 83.3  & 39.6 & 58.6 & 42.3 & 18.7 & 45.4 & 55.9 \\
		\checkmark & \checkmark & \checkmark &            & & 162.2 & 101.5 & 39.5 & 58.4 & 42.2 & 18.6 & 45.3 & 56.6 \\
		\checkmark &            & \checkmark & \checkmark & \checkmark & 136.0 & 72.5 & \textbf{40.1} & 59.2 & 42.7 & 19.4 & 45.7 & 57.1 \\

	\end{tabular}
	\label{tab:component}
\end{table}

\subsection{Ablation Study}\label{subsec:ablation}
We perform a thorough study of the design of FPG on COCO \texttt{val2017},
and explore different implementations of lateral connections within the grid. 
Our ablation experiments are conducted on RetinaNet with the \texttt{crop-aug} setting.

\noindent\textbf{Component Analysis.}
Firstly, we investigate the necessity of pyramid pathways and lateral connections.
Starting from a complete version of FPG with all connections and pathways,
we remove each component respectively to see the effects.
From Table~\ref{tab:component} we see that \emph{AcrossDown} (AD) is essential for FPG,
since it is the only connection that contribute to top-down pathways.
Removing these connections leads to a \textbf{$-$4.6} point mAP decrease.

On the contrary, \emph{AcrossUp} (AU) appears to be redundant, which only adds to more FLOPs and Parameters  but does not improve the performance.

Next, the connections \emph{SameUp} (SU) and \emph{AcrossSkip} (AS) are around equally beneficial, with a less severe impact on accuracy, as ignoring each of them results in a $-$0.5 mAP and $-$0.6 mAP decrement, respectively.

Finally, our grid contraction (Cont in Table~\ref{tab:component}) which truncates the lower-triangle low-level feature maps of the first 3 pathways (described in \S\ref{sec:grid_details} and illustrated in \figref{fig:fpg}) significantly reduces FLOPs and parameters, while maintaining the same level of performance. This shows that the lower, large-resolution features can use more shallow lateral structure, without sacrificing performance. Our hypothesis is that low-level feature maps need first to be enriched by top-down propagation before expanding into a deeper high-resolution pathway structure.

\begin{table}[h]
	\centering
			\vspace{-10pt}
	\caption{\small Comparison of different designs of \emph{SameUp}. \textbf{Bold}:  Default.}\label{tab:sameup-design}
	\vspace{2pt}
	\tablestyle{6pt}{1.05}
	\begin{tabular}{ l | l l | c c c | c c c}
		& FLOPs & Params & AP & AP$_\text{50}$ & AP$_{75}$ & AP$_\text{S}$ & AP$_\text{M}$ & AP$_{L}$ \\
		\shline
		AvgPool & 128.1 & 54.8 & 39.0 & 58.4 & 41.8 & 19.1 & 44.4 & 54.7 \\
		MaxPool & 128.1 & 54.8 & 39.6 & 58.8 & 42.6 & 18.8 & 45.3 & 55.9 \\
		\textbf{Conv}    & 136.0 & 72.5 & \textbf{40.1} & 59.2 & 42.7 & 19.4 & 45.7 & 57.1 \\
	\end{tabular}
			 \vspace{-10pt}
\end{table}

\noindent\textbf{SameUp ($\uparrow$).}
Table~\ref{tab:sameup-design} shows the ablation results of the SameUp connection in the pyramid pathway.
We compare three commonly used downsampling methods: average pooling, max pooling and $3\times 3$ convolution with a stride of 2.
Max pooling outperforms average pooling by +0.6 mAP which is further improved by using Conv (+0.5 mAP).

\begin{table}[h]
	\centering
	\caption{\small Comparison of different designs of \emph{AcrossSkip}. \textbf{Bold}:  Default.}
	\vspace{2pt}
	\tablestyle{6pt}{1.05}
	\begin{tabular}{ l | l l | c c c | c c c}
		& FLOPs & Params & AP & AP$_\text{50}$ & AP$_{75}$ & AP$_\text{S}$ & AP$_\text{M}$ & AP$_{L}$ \\
		\shline
		identity  & 133.0 & 70.2 & 39.5 & 58.1 & 42.3 & 19.1 & 45.3 & 56.2 \\
		\textbf{k1}        & 136.0 & 72.5 & \textbf{40.1} & 59.2 & 42.7 & 19.4 & 45.7 & 57.1 \\
	\end{tabular}
	\label{tab:skip-design}
\end{table}

\noindent\textbf{AcrossSkip (\xscolor{$\curvearrowright$}).}
Skip connection ease the training of deeper pyramid structures by propagating information across a direct connection path.
We compare two lightweight designs, an identity connection and $1\times 1$ convolutional projections.
As shown in Table~\ref{tab:skip-design}, $1\times 1$ convolution (k1) outperforms identity connection (identity) by +0.6 mAP with only marginal extra cost.

\begin{table}[h]
	\centering
	\caption{\small Comparison of different designs of \emph{AcrossDown}. \textbf{Bold}:  Default.}
	\vspace{2pt}
	\tablestyle{6pt}{1.05}
	\begin{tabular}{ l | l l | c c c | c c c}
		& FLOPs & Params & AP & AP$_\text{50}$ & AP$_{75}$ & AP$_\text{S}$ & AP$_\text{M}$ & AP$_{L}$ \\
		\shline
		intp      & 109.3 & 57.2 & 31.9 & 49.6 & 34.0 & 14.1 & 36.3 & 45.8 \\
		intp + k1 & 112.3 & 58.9 & 39.2 & 58.0 & 42.1 & 18.5 & 44.6 & 55.8 \\
		\textbf{intp + k3} & 136.0 & 72.5 & \textbf{40.1} & 59.2 & 42.7 & 19.4 & 45.7 & 57.1 \\
	\end{tabular}
	\label{tab:acrossdown-design}
\end{table}

\noindent\textbf{AcrossDown (\xdcolor{$\searrow$}).}
Finally,  we ablate  the structure of the across-pathway top-down connections.
The simplest implementation is nearest interpolation (intp), as used in FPN.
Hypothesizing that a na\"ive interpolation may not be enough to build strong top-down pathways,
we ablate either a $1\times 1$ (k1) or $3\times 3$ (k3) convolution to improve the FPGs capacity to project features for downsampling. 

Table~\ref{tab:acrossdown-design} shows the result in comparison. The accuracy is as low as 31.9 mAP with direct interpolation (intp). Adding an additional $1\times 1$ convolution (k1) improves it by \textbf{+7.3} mAP and adopting a larger kernel size (k3) leads to a further +0.9 mAP improvement.
Suggesting that a convolutional layer after interpolation, that can adapt the features and re-align the receptive field for further processing, is critical for the implementation of FPG and to achieve good performance.

\section{Conclusion}
\label{sec:conclusion}
This paper has presented Feature Pyramid Grids (FPG), a deep multi-pathway feature pyramid network, that represents the feature scale-space as a regular grid of parallel \mbox{pyramid} pathways. The pathways are intertwined by multi-directional \mbox{lateral} connections, forming a unified grid of feature pyramids. 
On instance detection and segmentation tasks, FPG provides significant improvements over both FPN and \mbox{NAS-FPN} with advantageous accuracy to computation trade-off. 
Given its unified and intuitive nature, we hope that FPG can serve as a strong component for future research and applications in instance-level recognition. 

\paragraph{Acknowledgments.}
We are grateful for discussions with Kaiming He and Ross Girshick. 

\newpage

\newcount\cvprrulercount
\appendix

\section*{Appendix}
\setcounter{table}{0}
\renewcommand{\thetable}{A.\arabic{table}}

\section{Qualitative Results} \label{app:qualitative_results}

As mentioned in the main paper, we show more qualitative results for comparing FPN\cite{lin2017feature}, NAS-FPN~\cite{ghiasi2019fpn}  and FPG here. In \figref{fig:more_ins_seg_example}, we observe that FPG is more accurate for predicting small-scale objects and partly occluded objects. For example, in the left column of \figref{fig:more_ins_seg_example}, it is seen that FPG is able to correctly predict masks for partly-occluded people who are spectating cross-country skiing, or in the second column of \figref{fig:more_ins_seg_example}, there are correct FPG predictions of `bench' instances in the background, while \eg FPN misclassifies these as `car' and NAS-FPN does not detect them.   We hypothesize that this is due to the deep feature pyramid representation of FPG which allows the network to build strong features for classifying small-scale objects in high-resolution features. 

\begin{figure}[htb]
	\includegraphics[width=\linewidth]{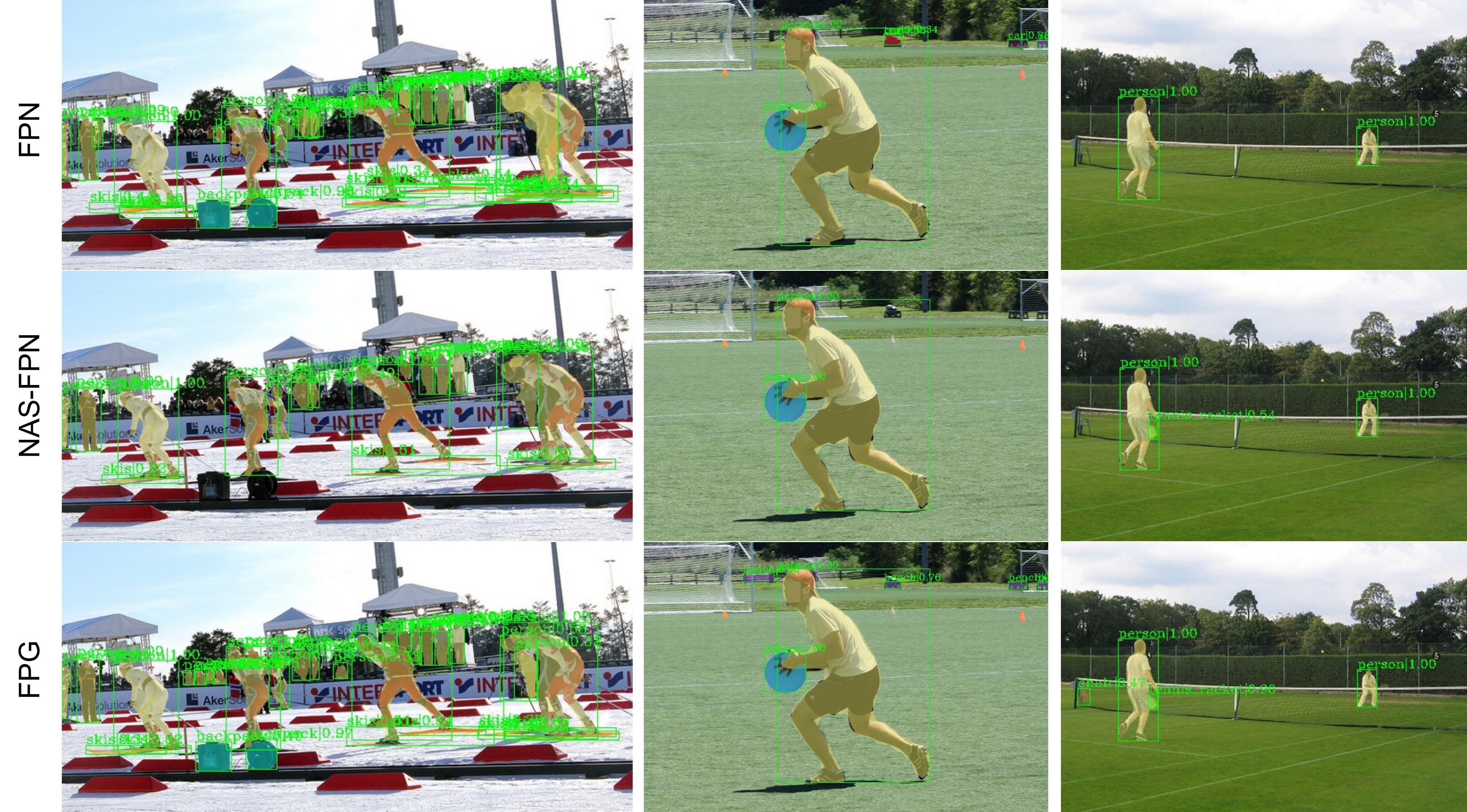}
\caption{\small \textbf{More examples for instance segmentation} with Mask R-CNN~\cite{he2017mask} using ResNet-50~\cite{He2016} with FPN~\cite{lin2017feature}, NAS-FPN~\cite{ghiasi2019fpn} and our FPG. Note how FPG is able to produce correct mask predictions for small-scale objects and has fewer misclassifications. Please view electronically, with zoom.}
	\vspace{-2mm}
	\label{fig:more_ins_seg_example}
\end{figure}

\section{Comparison with other pyramid networks} \label{app:comparison}
\begin{table}[ht]
	\fontsize{8pt}{1em}\selectfont
	\vspace{-15pt}
	\centering
	\caption{\textbf{Object detection} mAP based on Faster R-CNN and Mask R-CNN with different pyramids.}
	\begin{tabular}{ l | l | l | c c l | c c c c c c}
		Detector & Backbone & Pyramid & FLOPs & Params & AP & $\text{AP}_{50}$ & $\text{AP}_{75}$ & $\text{AP}_{S}$ & $\text{AP}_{M}$ & $\text{AP}_{L}$ \\
		\shline
		\multirowcell{5}{Faster\\ R-CNN } 
		& ResNet-50 & PA-FPN~\cite{liu2018path} & 117.1 & 52.2 & 37.7 & 58.4 & 40.8 & 18.4 & 40.6 & 50.8 \\
		& ResNet-50 & \textbf{FPG}   & 99.1  & 42.1 & \textbf{40.0(+2.3)} & 59.9 & 43.5 & 20.0 & 43.6 & 53.8 \\
		& HRNet-W18  & HR-FPN~\cite{sun2019deep}  & 83.6  & 27.5 & 37.5 & 57.7 & 40.9 & 19.4 & 39.7 & 49.8 \\
		& HRNet-W18  & \textbf{FPG}   & 91.2  & 28.0 & \textbf{39.4(+1.9)} & 59.3 & 42.9 & 20.9 & 41.8 & 51.4  \\
		\hline
		\multirowcell{5}{Mask\\ R-CNN} 
		& ResNet-50 & PA-FPN~\cite{liu2018path} & 185.6 & 54.8 & 38.2 & 58.5 & 41.5 & 17.8 & 41.2 & 52.7\\
		& ResNet-50 & \textbf{FPG}   & 161.8 & 44.4 & \textbf{40.9(+2.7)} & 60.5 & 44.6 & 20.9 & 44.4 & 54.6 \\
		& HRNet-W18  & HR-FPN~\cite{sun2019deep} & 152.1 & 30.1 & 38.4 & 58.4 & 41.8 & 19.6 & 40.7 & 50.7 \\
		& HRNet-W18  & \textbf{FPG}   & 153.9 & 30.3 & \textbf{40.3(+1.9)} & 59.8 & 44.0 & 21.5 & 42.9 & 52.4 \\
		\hline
	\end{tabular}
	\label{tab:results}
\end{table}

As referenced in the main paper, this section compares FPG 9@128 with related FPN structures and backbones from the literature: PANet~\cite{liu2018path} and HRNet~\cite{sun2019deep} on both Faster R-CNN and Mask R-CNN detectors. All experiments are conducted with the \texttt{crop-aug} setting, which is described in the implementation details, \sref{sec:exp_setup}. PANet~\cite{liu2018path} extends FPN with a path-aggregation pyramid structure (PA-FPN), and HRNet~\cite{sun2019deep} is a newly designed backbone that maintains high-resolution  through the whole feedforward process.
It achives better performance than ResNet backbones \cite{He2016} in several recognition tasks. Our FPG is aimed at better pyramidal feature representation and therefore could be complementary to HRNet backbone, if used instead of the pyramid in HRNet (HR-FPN). 

Results are shown in Table~\ref{tab:results}. We first compare to the pyramid of PANet~\cite{liu2018path}.  The table shows that \textbf{FPG} achieves +$\textbf{2.3}$ and +$\textbf{2.7}$ higher mAP than PA-FPN~\cite{liu2018path} on Faster R-CNN and Mask R-CNN, respectively, while being lighter in terms of Floating Point Operations (FLOPs) and parameters. This result is achieved under \textit{identical settings}, by just changing the feature pyramids of the detectors. 

We note the original PANet publication~\cite{liu2018path} reports higher performance by introducing extra components other than PA-FPN, such as adaptive pooling, fully connected fusion,  synchronized BN in the backbone, and heavier heads than the original ones used in R-CNN variants. These improvements to the  detection architectures are orthogonal to the FPN structure and expected to be complementary.  
For direct comparison, we only compare FPG to PA-FPN, \ie, the path aggregation feature pyramid structure, holding everything else constant.

Second, we evaluate replacing the feature pyramid used in HRNet~\cite{sun2019deep} with FPG (or equivalently, changing the backbone of FPG from ResNet to HRNet) in Table~\ref{tab:results}.
HRNet-W18 + \textbf{FPG} improves box AP by \textbf{+1.9} over HRNet-W18 + HR-FPN on both Faster R-CNN and Mask R-CNN, showing complementary of FPG with the underlying backbones used for detection and superiority to the default HR-FPN under similar FLOPs and parameters.

\clearpage
%
%
\bibliographystyle{splncs04}
\bibliography{fpg}
\end{document}